# On solving decision and risk management problems subject to uncertainty

Alexander Gutfraind, PhD
agutfraind.research@gmail.com

## Abstract

Uncertainty is a pervasive challenge in decision and risk management and it is usually studied by quantification and modeling. Interestingly, engineers and other decision makers usually manage uncertainty with strategies such as incorporating robustness, or by employing decision heuristics. The focus of this paper is then to develop a systematic understanding of such strategies, determine their range of application, and develop a framework to better employ them.

Based on a review of a dataset of 100 decision problems, this paper found that many decision problems have pivotal properties, i.e. properties that enable solution strategies, and finds 14 such properties. Therefore, an analyst can first find these properties in a given problem, and then utilize the strategies they enable. Multi-objective optimization methods could be used to make investment decisions quantitatively. The analytical complexity of decision problems can also be scored by evaluating how many of the pivotal properties are available. Overall, we find that in the light of pivotal properties, complex problems under uncertainty frequently appear surprisingly tractable.

# Introduction

Resolving most decision or risk management problems requires addressing uncertainty in all its forms (Yoe 2019; Schultz et al. 2010). While uncertainty can be modeled and managed using probability theory (Cox 1946; Savage 1951; Hacking 2006), probabilistic methods have come under a sustained attack since at least the 1920s in the work of Knight (Knight 2014) and Keynes (Keynes 1990). To the critics, the issues are: (1) It is impractical to build a forecasting model for the many outcomes of interest in the complex problems found in engineering, finance and other fields (Kay and King 2020; Sniedovich 2012); (2) Some problems are often too ambiguous or under-determined to be quantified using a model (Colyvan 2008; Regan, Colyvan, and Burgman 2002); and (3) If a model could be built, it would depend on parameters that cannot be estimated from data or past experience (Kay and King 2020; Gerrard and Tsanakas 2011) and in particular would involve low-probability, high-consequence events that are affected by high uncertainty (Waller 2013).

In aggregate, the critics argue that many situations involve uncertainty that is too difficult to quantify and requires special strategies (for responses see e.g. (Friedman 1976; Waller 2013; Aven 2013)). Various terms have been offered in the literature to describe problems that are resistant to quantitative methods, including "Knightian uncertainty", "Hard uncertainty" (Vercelli 1998), "Unknown Unknowns" (Rumsfeld 2011), "Black Swans" (Taleb 2007) and "Radical uncertainty" (Kay and King 2020), all with broadly similar meanings. However, these terms are quite difficult to operationalize, and the best-known cases of unknown unknowns are disputed (Aven 2013; Ale, Hartford, and Slater 2020). Seemingly impossible forecasting problems, such as forecasting sudden geopolitical events and financial crises, have actually been anticipated (Wucker 2016).

This article's primary contribution to the literature is to offer a new framework to discuss problems under uncertainty. It argues that decision problems, even when affected by high uncertainty, often have pivotal properties (defined and listed below). It points out that the analytical hardness of a problem could be estimated by looking at the presence or absence of these properties. Pivotal properties are then linked to strategies for risk management and decision making that do not depend on quantification of uncertainty. Optimal investment in such strategies could be determined using multi-objective optimization or heuristically, as opposed to more heuristic solutions proposed to date (Sniedovich 2012; Taleb 2008; Kay and King 2020).

The project of forming a taxonomy of uncertainty has an extensive literature, notably in foundational categories such as Knight's distinction (Knight 2014), Hacking's identifying aleatory and epistemic uncertainty (Hacking 2006), and the distinction of uncertainty from ambiguity (Colyvan 2008). Recently, there have been domain-specific taxonomies of uncertainty (Regan, Colyvan, and Burgman 2002; Ristovski et al. 2014). Lo and Mueller (2010) identify five levels of progressively high uncertainty. Concurrently, the Computer Science community developed a rich theory of computational complexity(Arora and Barak 2009), but this theory only considers problems that could be defined in the strict context of Turing machines or similar computational devices, rather than the complex societal context considered by decision theory and its practitioners. The idea of subjective probability (Savage 1951; Ramsey 2013; Bier and Lin 2013) seems to imply that the hardness of a problem is observer-dependent, but some argue that subjective probability cannot be used effectively in a large class of problems(Kay and King 2020). To overcome this uncertainty about uncertainty, the scoring rule proposed here is derived from the properties of the decision problem and therefore could be applied objectively to a given problem.

## Two motivating examples of uncertainty

To motivate the development of pivotal properties, we will review several examples of uncertainty from the large list of scenarios underlying this study. The first example is the problem originally considered by Knight and Keynes of entrepreneurial investment, reviewed in Sakai 2019. An entrepreneur is seeking to found a new firm and the goal is to maximize the future value of the investment. If the business model or technology is new, there is little quantitative basis for comparing actions and forecasting outcomes. It is impossible to forecast key determinants of profitability, e.g. whether the market will see value in the new technology, whether the firm will be well-managed, or whether the economic and financial conditions will be favorable over the next 10 years. General benchmarks and trends are of course available for such investments, but they suffer from high dispersion and provide little specific guidance for choosing investments. For a second example, consider the problem of a health department preparing for a pandemic. Pandemics are rare, hard to predict and the biology of the pathogen varies, resulting in varying impact and the required mitigation strategies.

Contrasting the disparate problems in these scenarios reveals two common general properties. While investment decisions are nearly entirely based on future capital return (a single objective), pandemic decisions must weigh multiple objectives (e.g. mortality, morbidity, economic impacts) and multiple stakeholders (e.g. different demographic groups). Secondly, risk in financial decisions can often be transferred to another party – for example, by buying an annuity contract with an insurance company, or pooling resources with another investor. In pandemic preparedness, except for some financial and resource risks, the major risks to health cannot be transferred.

Yet the two scenarios are similar in some surprising ways. Uncertainty in both scenarios could be reduced through the acquisition of knowledge, e.g. fundamental research in either the technology or the biology of pathogens, or more tactically, scenario planning or computational modeling. Secondly, the decision makers have a degree of control over the system they are responsible for. The entrepreneur has large discretion over decisions such as hiring of experts, the R&D roadmap and marketing. To a lesser extent, the local health department has discretion over the drawing of plans, training of healthcare staff, and acquisition of resources such as masks. In both cases, there is also a large world outside of their control that they need to consider (e.g. consumers and capital markets; the at-risk population and other public health authorities). Additionally, in what could be called indexability, the outcomes in both situations could be expressed adequately using random variables (e.g. the market value of the investment, the number of infections and a few others). Indexability should be distinguished from quantification (e.g. the ability to model or predict): the price of a company's stock indexes the value of the investment but does not eliminate the problem of economic forecasting or black swan events.

The above pivotal properties of the problems lead to strategies to reduce the uncertainty or even the ability to make a decision. Whenever outcomes could be improved by more data, we can invest resources in gathering data or fundamental knowledge. This property should not be taken for granted because in some situations we lack a clear guide as to where our knowledge gaps are (true unknown unknowns), or we might lack the ability to carry out a study since the system is too complex, inaccessible with current methods, or not subject to experimentation, as occurs in some complex systems.

Second, control over system design enables many strategies, and it is the default framework in engineering. The basic strategy is to make the design "robust"; that is, to give the design the strength to withstand unexpected events. This includes specific design tactics such as "defense in depth" and redundancy. Outside of engineering, the designer often has partial control over certain system components, leading to new strategies. For example, in financial portfolios the engineer has control over the components of a portfolio or trading strategy, allowing him to reduce exposure to certain risks, or make the portfolio resilient by adding financial hedges.

The ability to precisely define outcomes (even if they cannot be forecasted) can be highly valuable for risk management. It may enable estimation of volatility, setting of benchmarks, and trading of risk with counterparties. It enables interactive trial-and-error approach to decisions until the outcome is improved (even if the system is difficult to study or model). When indexes are hard to produce (e.g., when the outcomes are fuzzy or impractical to measure), all of the above strategies become inaccessible.

# Pivotal properties

Generalizing from the above examples, we are interested in characterizing a broad range of problems affected by a high degree of uncertainty, and identifying properties that could aid in their classification and resolution. The problems of interest are all problems studied in decision theory and risk analysis: individual choices and large-scale societal problems; decisions with stakes both low and high. The goal of the problem might be optimization of an outcome subject to uncertainty, or reduction of risk where the risk is difficult to quantify.

To be a little more precise, we will adopt the states-actions-outcomes setting that's universal to decision theory and risk analysis (Borgonovo et al 2018). The decision makers face an uncertain current state of the world, $s \in S$; have to choose between alternative actions from a set, $a \in F$; and experience future outcomes (also called objectives): $y = F(w, a) \in O$. Our decision makers are often unaware of the possible actions and outcomes either at onset or later (Steele and Orri Stefansson 2021), unlike in classical settings. The action set might be very small (e.g. choosing between two options) or very large (as happens in problems of system design). In most cases we will think of the action as occurring at a single point in time, but in certain situations we face multi-stage problems, as well as an infinite sequence of decisions. Solving a problem here usually means finding a satisficing solution in the sense of Simon (Artinger, Gigerenzer, and Jacobs 2022), and only in special cases the solution would be optimal in any sense of the word. Concretely, solving might mean optimizing (as in classical decision theory) but more often could mean finding a feasible (constraint-satisfying) action, or merely not causing harm to our objectives.

## Identification of common properties

In order to identify properties of problems, the following methodology was employed. First, the author and several paid assistants collected examples of problems from various domains, and recorded strategies used by practitioners for decision making and risk mitigation. Then, pivotal properties of the problem were found by (a) contrasting the problems to each other, (b) identifying properties of the problem that make solution strategies possible, and (c) the problem-specific properties of problems were unified names. The list thus created cannot ever include all pivotal properties of all problems because problems are virtually infinite in variety, and many problems have unique useful aspects. Fortunately, it should give commonly-found properties of many problems, and allow them to be classified and scored. The strategies decision-makers named tended to be heuristics that did not guarantee finding optimality, but as a consequence they did not depend on quantification, and therefore were inherently robust to uncertainty affecting models.

## Properties of problems and important classes of strategies

Analysis of the decision scenarios produced 14 pivotal properties (Table 1). The properties might be loosely clustered to (1) those dealing with the decision context and (2) those dealing with the action and event spaces.

**Table 1**: Common properties of decision problems and solution strategies robust to uncertainty. See Key below for definitions of properties. Strategies may fit under multiple properties or require several conditions.

| # | Property of problems | Robust solution strategies |
|---|---|---|
| 1 | Easily satisficeable | Find a feasible solution, Use the default action, Select a solution using heuristics |
| 2 | Reversible with acceptable losses | Best guess, trial-and-error, Wait and see |
| 3 | Few objectives and/or Few similar stakeholders | Optimization of total outcome; Decision by polling or delegation (while maintaining fairness) |
| 4 | Delayed onset, or drawn-out impact | Expansive analysis, Preparation and planning |
| 5 | Small event space, or Small action space | Evaluation and comparison of all events and actions, Focused planning |
| 6 | Controllable system design | Robust design, Resilience planning, Dedicated response unit, Optionality, Evolutionary architecture |
| 7 | Indexable outcomes | Statistical modeling, Computation of benchmarks and volatility |
| 8 | Transferable risk | Risk contracts, Insurance & financial instruments |
| 9 | Learnable phenomenon | Basic research, Meta-learning of unknowns, Knowledge dissemination |
| 10 | Well-understood phenomenon | System models (mathematical, computational, predictive), Maximization of expected utility, Probabilistic risk management, Decision templates, Expert elicitation and judgment |
| 11 | Well-understood adversary | Randomization and game-theoretic analysis, Misdirection/Deception, Rapid adaptation |
| 12 | Sequentially interactable system | Trial-and-error, Stochastic search, Reinforcement learning |
| 13 | Detectable hazard | Early detection before impact, Rapid identification after event |
| 14 | Bounded hazard | Minimize area or time of impact, Deflect, Delay |

<u>Key to the table:</u> Easily satisficeable refers to problems where feasible solutions are easily found and/or where many of the feasible solutions are likely acceptable to the stakeholders, and therefore finding and comparing alternative solutions is easier. Reversible means that the action could be changed in the future with acceptable costs. Few objectives means that only one or several objectives are important, and few or similar stakeholders means that the action needs to satisfy only a small number of stakeholders, or equivalently a group that's internally similar to each other. Small event or action space means that the set of possible actions or outcomes is small. Delayed onset or drawn-out impact refers to events that are anticipated to occur after a relatively long period of time, or to have gradual (and thus potentially reducible) impact. Controllable system design means that it is possible to design the system at risk for the event, or at least a subset of that system (see below). Indexable outcomes are outcomes that could be expressed using random variables (e.g. the market value of the investment, the number of infections and few others). Transferable risk means that the risk could be transferred substantially to another party. Learnable phenomenon refers to whether additional research and analysis is productive, and often means that the phenomenon is bound by laws or processes (possibly with time-varying parameters). Well-understood phenomenon means that a body of knowledge exists to describe the phenomenon such as scientific understanding, data or human experience. Well-understood adversary means that the decision must consider an adversary that acts to improve their outcome and can reduce ours, but we have knowledge about their capabilities. Sequentially interactable refers to the ability to perform actions and observe outcomes, gradually learning the system and to optimize actions. Detectable means that the event could be recognized before it occurs, or soon after it has occurred. Bounded hazard means that the event is bounded by space, time or other measures (e.g. environmental contamination, extreme weather), as opposed to having an essentially total effect, or affects all of our outcomes or objectives.

A few words should be said about the Controllable system design property - the ability to design the system at risk. It is one of the most useful properties in the list, since it unlocks a large array of strategies. In various areas of engineering, the mainstay solution to uncertainty is robustness - constructing with a factor of safety, minimizing and hardening the attack surface, implementing sacrificial parts, defense-in-depth, compartmentalization and others. In financial engineering, if one has control over the system at risk (e.g. an asset portfolio), one could implement diversification, minimization of exposure to certain risks, and other techniques. Lastly, one frequently has no control over the system as a whole, but a budget to design the response - to stand up a dedicated risk management or response team, increasing resilience, implementing training and more. In some cases the risk is positive (e.g. sudden profit, introduction of new technical capabilities). Positive outcomes could be harvested by designing for optionality and evolutionary architecture.

## Using pivotal properties in practice

Decision makers attempting to use this framework should start by developing a detailed understanding of the problem, the decision makers and the context. Next, the pivotal properties of the problem should be listed based on Table 1. The properties should then immediately produce a list of possible solution strategies from Table 1. The problem might also have additional pivotal properties that lead to special solution strategies. Next, strategies listed in Table 1 should be evaluated and defined as specific possible actions.

Selecting actions under conditions of uncertainty requires a new strategy. While in risk management it is generally assumed that we can calculate the costs and benefits of actions, this is not true in general. Due to uncertainty about outcomes, it is neither possible to use expected utility to choose between actions nor to decide how much to invest in a given action.

There are both practical and quantitative solutions to this challenge. In practical cases, decision makers often make decisions based on criteria such as subjective confidence and past experience, heuristics and analogical cases, and external criteria (rules, standards, or guidelines).

A quantitative solution is often possible as well, as follows. Suppose without loss of generality that our problem is focused on maximizing certain system performance indicators subject to constraints:

$$\max_{x}\{K_1(x),..,K_p(x)\} \text{ such that } P(x) \leq Q. \qquad (1)$$

Even under uncertainty, it is often possible to quantify how actions benefit certain metrics of interest $L$ (e.g., resilience, knowledge), even if they cannot be mapped to primary outcomes (e.g., future yield of an investment). Indeed, there is a fairly elaborate mathematical theory around many of the approaches used in engineering and finance. For example, in resilience-based approaches, it is possible to quantify the system's resilience and apply computational optimization to increase the resilience (Ganin et al. 2016) without having to evaluate the effect on risk directly. Physical theory might be expressed in numerical differential equations, while detection strategies might be supported by machine learning. Indeed, even methods that rely on expert judgment could be systematized and quantified in order to reduce uncertainty and bias.

Therefore, we let the relationship between action and the outcomes be the mapping $x \to \{L_1(x),..,L_q(x)\}$. To decide between outcomes we formulate an extended multi-objective problem:

$$\max_{x}\{K_1(x),..,K_p(x),L_1(x),..,L_q(x)\} \text{ such that } P'(x) \leq Q' \qquad (2)$$

The decision maker could then use multi-objective optimization to eliminate dominated solutions and find the best trade-offs between different outcomes. For illustration, the decision maker might invest some amount in resilience or basic knowledge up to a point of rapidly diminishing returns, even though it is impossible to quantify how these investments would improve the future yield of an investment.

## Scoring complexity of problems

An application of pivotal properties is to provide a score of the analytical complexity or difficulty of problems. Generally, the scoring function should give higher values of complexity when the problem lacks a property, or when the solution strategies cannot be applied. Concretely, suppose that for problem $z$ the function $R_i(z) \in [0,1]$ gives the extent to which a property $i$ resolves the problem, in the sense that removing this property, while keeping everything else unchanged, would decrease the difficulty of the problem. If the property does not apply to the problem, then $R_i = 0$.

Analytical complexity could be defined as the product:

$$H(z) = \prod_i (1 - R_i(z)) \qquad (3)$$

In general, we will need a separate resolution function for each property (e.g., for the count of possible events, hyperbolic tangent of the number of possible events). Using the product function in Eqn. (3) reflects the effect that the complexity of the entire decision problem can become 0 if $R_i = 1$ for any $i$. For example, any decision problem is analytically trivial if there is only one possible action. Additionally, the product function guarantees that finding additional properties would not affect $H(z)$ if they do not assist in resolving the problem. The simplest version of this definition is to make $R_i \in \{0, c\}$ for some positive constant $c \in (0,1]$, i.e., just indicate the absence or presence of the pivotal property in problem $z$. Then the complexity of the problem simplifies to $H(z) = (1 - c)^k$, where $k$ is the number of pivotal properties.

The advantage of such scoring is that it helps us move away from dichotomous and observer-dependent terms such as the "unknown unknown": once we specify the analyst facing the problem, we can compute the complexity, that is to say, the complexity would be constant for all reasonably informed analysts at a given point of time. The framework accounts for the critique of subjective probability, namely, that some observers will be much more informed than others and possess precise subjective probabilities[1], but it rejects the idea that the existence of precise subjective probabilities destroys the possibility of analytical complexity - if some observers are much more informed than others, then the score would be observer-class dependent. In any case, of the many properties in Table 1, only a few are epistemic (well-known and learnable). Whether the complexity is affected by some property $i$ depends on what analytics tools are available to exploit this property, and could potentially change through methodological advancement.

The resulting score is somewhat different from existing hierarchies of uncertainty in that it quantifies the complexity of the problem, rather than of the underlying uncertainty, and this is desirable. While many problems are affected by profound uncertainty, they are not hard from a decision theory point of view. For example, long-term investment decisions are classical examples of Knightian uncertainty, but they are easy if the number of choices is small. Many investment advisors merely propose that clients choose whether to invest 60% of their portfolio in an index of stocks and the rest in bonds, or vice versa. In our analysis of the empirical dataset described in Methods, in every problem we could always find some pivotal properties and mitigation strategies. One may hypothesize that the majority of problems that are affected by large uncertainty and are hard to address with expected utility methods are nevertheless solvable. A hard problem that has become a major public concern - an unexpected planetary disaster, such as an asteroid impact (Reinhardt et al. 2016) - has many pivotal properties: humanity has a limited set of objectives (foremost, survival), the scenario space is relatively small, the event it detectable, stationary and learnable, and so on.

This is not to deny that there might exist situations, the hardest nuts, that have few pivotal properties and consequently cannot be managed with any of the strategies. We can imagine such a scenario by negating the pivotal properties; e.g., poorly understood, hard to learn, non-stationary, not detectable in advance, lack of system design or control over response. The best examples of hard problems are perhaps those involving complex adaptive systems that are connected to human society in complex ways. Examples therefore include environmental and economic problems - the well-known area of "wicked" planning problems (Rittel and Webber 2018).

---

[1] For example, ancient Egyptian astronomers could reputedly forecast the annual flood of the Nile river. Therefore, although they did not have the machinery of probability theory, they associated less uncertainty with the flood event than surrounding nations who viewed the event as highly unpredictable.

## Discussion and Conclusions

This study is a reexamination of uncertainty in decision theory and risk. It is inspired by the paradoxical situation where, on the one hand, virtually all problems are profoundly affected by uncertainty, and yet practitioners seem to have a large arsenal of solution strategies. The paradox is resolved by noting that problems have properties that provide a backdoor to finding good solutions. These properties allow us to quantify the analytical complexity of problems.

More broadly, the toolkit of pivotal properties makes a small step towards better incorporating the problem of profound uncertainty within the discipline of risk management. Responding to critics of modeling techniques, this article proposes that decision problems usually have pivotal properties that offer a new method to describe the analytical complexity of problems. Additionally, they enable quantitative methods for decision analysis and risk management that could be applied across different domains such as finance, engineering and public policy.

## Acknowledgements

I would like to thank Profs. Vicki Bier, Michael Genkin and several colleagues for the helpful discussion that inspired and refined this study.